\pdfoutput=1

\documentclass[11pt]{article}

\usepackage[]{ACL2023}

\usepackage{times}
\usepackage{latexsym}

\usepackage[T1]{fontenc}

\usepackage[utf8]{inputenc}

\usepackage{microtype}

\usepackage{inconsolata}
\usepackage{amsfonts,amssymb}
\usepackage{mathtools}
\usepackage{booktabs}
\usepackage{arydshln}
\usepackage{multirow}

\title{Entailment as Robust Self-Learner}

\author{Jiaxin Ge$^{1*}$ \and Hongyin Luo$^{2}$\thanks{$\;\;$Equal contribution. Correspondence to Hongyin Luo at \texttt{hyluo@mit.edu}. Code and processed data are available at \url{https://github.com/luohongyin/EntST}.}$\;$ \and Yoon Kim$^2$ \and James Glass$^2$ \\
        $^1$ Peking University, Beijing, China \\ $^2$ MIT Computer Science and Artificial Intelligence Lab, Cambridge MA, US \\ \texttt{aomaru@stu.pku.edu.cn, \{hyluo, yoonkim, glass\}@mit.edu}}

\begin{document}

\maketitle
\begin{abstract}
Entailment has been recognized as an important metric for evaluating natural language understanding (NLU) models, and recent studies have found that entailment pretraining benefits weakly supervised fine-tuning. In this work, we design a prompting strategy that formulates a number of different NLU tasks as contextual entailment. This approach improves the zero-shot adaptation of pretrained entailment models. Secondly, we notice that self-training entailment-based models with unlabeled data can significantly improve the adaptation performance on downstream tasks. To achieve more stable improvement, we propose the \textbf{Sim}ple \textbf{P}seudo-\textbf{L}abel \textbf{E}diting (SimPLE) algorithm for better pseudo-labeling quality in self-training. We also found that both pretrained entailment-based models and the self-trained models are robust against adversarial evaluation data. Experiments on binary and multi-class classification tasks show that SimPLE leads to more robust self-training results, indicating that the self-trained entailment models are more efficient and trustworthy than large language models on language understanding tasks. 
\end{abstract}

\section{Introduction}
Although achieving state-of-the-art performance in different natural language understanding (NLU) tasks \cite{devlin2018bert,liu2019roberta,yang2019xlnet,clark2020electra,he2020deberta,joshi2020spanbert}, large-scale pretrained language models still highly depend on human-annotated, task-specific training corpora for fine-tuning because the self-supervised pretraining objective does not incorporate explicit task-related knowledge. As a result,  state-of-the-art language models are still challenged by the lack of adequate fine-tuning data and  difficult evaluation examples crafted by adversarial attacks or model-in-loop adversarial data annotations \cite{wang2021adversarial,bartolo2020beat,zang2019word,garg2020bae,li2020bert}.

On the other hand, entailment is recognized as a minimal requirement for NLU \cite{condoravdi2003entailment}. Recent studies have found that entailment learning improves sentence representation \cite{reimers2019sentence,gao2021simcse}. However, these models still need fine-tuning with human-annotated training  data to handle downstream NLU tasks. The authors of \citet{wang2021entailment} found that entailment-based models are also few-shot learners that outperform recent efforts on few-shot NLU. For example, LM-BFF \cite{gao2020making} proves that entailment learning can significantly improve the data efficiency and adaptation ability of language models.

In this work, we further explore the zero-shot and unsupervised adaptation abilities of entailment-based models without any human-labeled training corpora on downstream tasks. We first study the zero-shot and unsupervised adaptation abilities of the entailment-based language models. Inspired by recent progress on prompt tuning, we formulate different NLU tasks as contextual entailment \cite{routley1973semantics} by constructing task-specific suppositions. The language models are trained to predict the truth value of the constructed suppositions. In zero-shot adaptation experiments, we find this approach significantly outperforms naively concatenating different inputs and labels, proving that the supposition construction method mitigates the distribution gap among different NLU tasks.

We further explore the potential of the unsupervised adaptation ability of entailment-based models. We use the pretrained entailment models to predict the pseudo-labels of unlabeled, task-specific language data. We find that the entailment-based models can be improved with self-training \cite{blum1998combining} with the automatically annotated pseudo-labels \cite{he2019revisiting}. While the self-training strategy has been proven effective on different tasks and modalities \cite{zou2019confidence,zoph2020rethinking,meng2020text,xie2020self}, a major challenge for self-training is the unstable performance caused by the noisy pseudo-labels. A number of solutions have been proposed to mitigate this issue. The most popular methods are training data selection \cite{li2005setred,lang2022co} and pseudo-label editing \cite{shin2020two,mandal2020novel}. Recent work also found that simple Dropout \cite{srivastava2014dropout} approaches improve contrastive learning \cite{gao2021simcse} and speech recognition \cite{khurana2021unsupervised,dawalatabad2022unsupervised}.

To combine the benefits of data selection and label editing methods, we propose SimPLE, a \textbf{sim}ple \textbf{p}seudo-\textbf{l}abel \textbf{e}ditting algorithm with simple text augmentation, uncertainty-based data filtering, and majority-based pseudo-labeling. Experiments with different backbone models on binary, multi-class, regular, and adversarial NLU tasks show that our approach makes the following contributions,
\begin{itemize} \setlength{\itemsep}{0pt} \setlength{\parsep}{0pt}
\item Supposition-based task formulation improves the zero-shot adaptation and robustness against adversarial evaluation data of entailment models across different NLU tasks.
\item SimPLE improves the pseudo-labeling accuracy on confident and uncertain training samples, leading to significant improvement over all self-training and pretrained baselines.
\item Self-trained, 350M-parameter entailment models without human-generated labels outperform supervised language models with 137B parameters, proving the data and computation efficiency of entailment self-training.
\end{itemize}

\section{Related Work}
\label{sec:bg}
\textbf{Language modeling.} Task-agnostic, large-scale language models can solve a number of natural language understanding (NLU) tasks \cite{brown2020language,raffel2020exploring,lewis2019bart,wei2022emergent,wei2022chain}. On the other hand, pretraining with annotated training corpora of different natural language tasks also benefits the generalize ability and zero-shot adaptation performance \cite{sanh2021multitask}. Recent studies have found that textual entailment \cite{bowman2015large,N18-1101} is a powerful pretraining task. Entailment models are applied for sentence representation learning \citep{reimers-gurevych-2019-sentence,gao2021simcse}, relation extraction \citep{obamuyide2018zero,yin-etal-2019-benchmarking}, and fact-checking \citep{thorne2018automated}. The authors of \citet{wang2021entailment} showed that entailment models can benefit the few-shot learning performance of pretrained language models on NLU tasks.

\noindent \textbf{Robustness in Self-training.} While most self-training studies are under the computer vision context \cite{zoph2020rethinking,zou2019confidence}, efforts also exist for self-training the latest neural language models, including back translation \cite{he2019revisiting}, text augmentation \citep{xie2020unsupervised,chen-etal-2020-mixtext}, question-answer synthesis \cite{bartolo-etal-2021-improving,luo-etal-2022-cooperative}, and co-training \cite{lang2022co}. However, self-training methods suffer from noisy pseudo-labels. In computer vision, a straightforward solution is obtaining confident pseudo-labels by augmenting input images \cite{shin2020two,mandal2020novel,sohn2020fixmatch}, including shifting, rotating, or adding noise to pixels. However, data augmentation is not as straightforward for natural language if no additional model is used. Instead, some model-level methods can be applied. \citet{zou2019confidence} proposed regularizing over pseudo-label confidence to avoid overfitting to simple cases, \citet{gao2021simcse,khurana2021unsupervised} applied dropout to improve the quality of training corpora. \citet{li2005setred,lang2022co} applied a graph-based confidence estimation method for removing training samples with uncertain pseudo labels.

\noindent \textbf{Difference with previous work.} Without any additional language model for text augmentation, we propose a model-level, augmented pseudo-labeling method that improves self-training performance for entailment models. Our method avoids dropping training data and performs more stably than dropout-based methods. Different from previous work on weakly-supervised language understanding with entailment models \cite{wang2021entailment}, we do not use any human-generated labels. Our models contain 1/500 trainable parameters compared to the models used in \citet{lang2022co,sanh2021multitask}.

\section{Entailment Self-training}
\label{sec:entst}
\noindent \textbf{Pretraining.} Recent studies have found that entailment-based language models can efficiently adapt to different natural language understanding (NLU) tasks with a limited number of human-labeled training samples \cite{wang2021entailment,luo-glass-2023-logic}. In this work, we find that entailment models can be self-improved without any human-generated labels by constructing suppositions (prompts) that describe the given tasks. Most NLU tasks can be formulated as predicting the truth value of the constructed suppositions that wrap inputs and label descriptions, as shown in Table \ref{tab:sup}.
\begin{table}[h]
\small
\centering
\begin{tabular}{@{}lll@{}}
\toprule
\multicolumn{1}{c}{\textbf{Task}} & \multicolumn{1}{c}{\textbf{Inputs}} & \multicolumn{1}{c}{\textbf{Supposition}} \\ \midrule
MNLI  & \{p, h\}    & h is entailed by p.   \\
RTE   & \{p, h\}    & h is entailed by p.   \\
QNLI  & \{t, q\}    & The answer to q is entailed by t.    \\
QQP   & \{$q_1$, $q_2$\}      & $q_1$'s answer is entailed by $q_2$'s answer.  \\
SST2  & \{x\}       & The movie is good is entailed by x. \\
\bottomrule
\end{tabular}
\caption{The suppositions constructed based on the definitions of different GLUE tasks \cite{wang2018glue}.}
\label{tab:sup}
\end{table}

By training the entailment model using the MNLI corpus given with the constructed suppositions, the model can be directly adapted to other tasks with relatively high accuracy. We will show that without entailment pretraining, similar performance can only be achieved by 400 times bigger language models. The entailment-based models can be further fine-tuned on unlabeled texts via self-training. We apply different adaptation strategies for binary and multi-class classification tasks.

\noindent \textbf{Binary classification.} Supposition-based entailment models predict True, Neutral, and False scores for each supposition, corresponding to entail, neutral, and contradictory labels of the MNLI corpus. For binary classification, we ignore the neutral score and calculate only True and False probabilities, and the True/False predicted can be linked to corresponding labels according to the supposition. For example, the SST2 supposition in Table \ref{tab:sup} being true means that \{x\} is a positive movie review. The predicted True/False values are used as pseudo-labels for self-training,

\noindent \textbf{Multi-class classification.}
In binary classification, the model is presented with a single supposition and asked to decide whether it's true or not. In multi-class classification, the model is presented with a context sentence and multiple labels and is asked to choose the correct label. 

To predict the correct answer from multiple options, we propose an entailment score ranking method.
 First, for each sentence to be classified, we construct a supposition for each label. For example, in an emotion classification task, given the sentence \texttt{S}, we construct the following suppositions: "I am happy is entailed by \texttt{S}", "I am sad is entailed by \texttt{S}", and "I am shocked is entailed by \texttt{S}". We calculate the entailment probability of each supposition with the entailment model and predict the label associated with the most entailed supposition.

 We propose a max-confidence tuning method for self-training.
 We select the class with the highest entailment score and then record its predicted pseudo-label for further self-training, and ignore other classes. The model does not need to classify each class correctly but merely learns to predict the truth value of its most confident supposition.

\begin{figure*}[h]
\centering
\includegraphics[width=\textwidth]{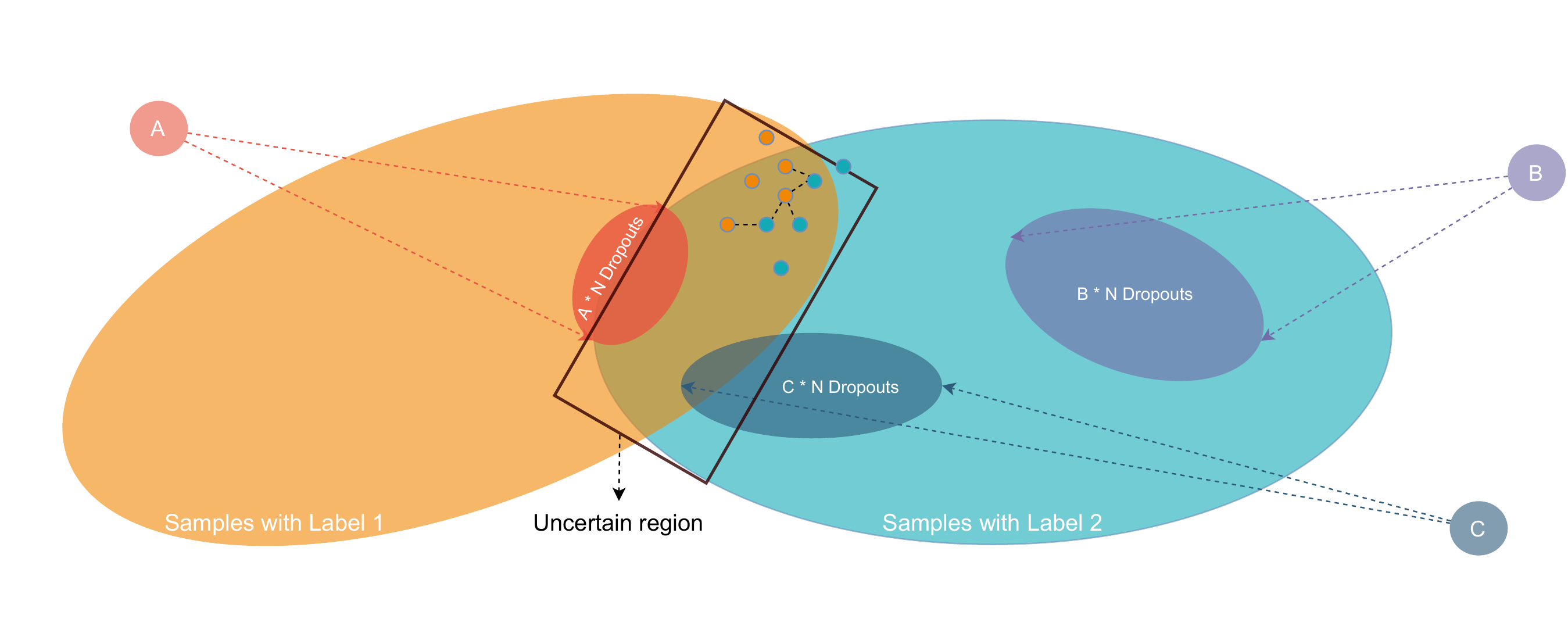}
\caption{Visualization of the SimPLE method. The figure shows the embedding space of natural sentences, and different colors represent different predicted labels. Each data sample is labeled with multiple random dropouts, and we use the SETRED algorithm to detect the uncertain pseudo-labels. The final label is voted by confident inferences.}
\label{fig:pipeline}
\end{figure*}

\section{Simple Pseudo-label Editing}
\label{sec:simple}
We propose the simple pseudo-label editing (SimPLE) method, a three-step pipeline for generating robust pseudo labels, including augmented pseudo-labeling with dropout, uncertain data filtering, and majority-based relabeling. We introduce the details of each step in this section.

\subsection{Simple Augmentation for Pseudo-labeling}
Because of languages' discrete and sequential nature, changing a token in a sentence might completely invert its meaning. As a result, unlike straightforward and effective image augmentation processes like FixMatch \cite{sohn2020fixmatch}, additional augmentation models are usually needed for text augmentation. Recent studies have found that instead of data-level augmentation, the Dropout mechanism leads to decent embedding-level augmentation. \citet{gao2021simcse} applied dropout for contrastive sentence representation learning, and \citet{khurana2021unsupervised} selected confident pseudo-labels by measuring the consistency of a model with the same input data and random dropouts.

As the first step of generating augmented pseudo labels, we run $N$ independent evaluations with random dropout (dropout rate $=0.1$) for each input training sample $x_i$ and obtain a set of $N$ noisy pseudo-labels.
\begin{equation}
Y_i = \{y_j = M_j^*(x_i)\; |\; j \in [0, N)\}
\end{equation}
where $j$ stands for the $j$-th independent evaluation with a dropout model $M^*$. Meanwhile, we store a set of sequence representations $E_i = \{e_0, e_1, \dots, e_{N-1}\}$ of $x_i$ collected in each feed-forward process. After finishing this step, we collect a set of data, pseudo-label, and embeddings.
\begin{equation}
\label{eq:c}
C = \{(x_i, y_i^j, e_i^j) \; | \; i \in [0, M), j \in [0, N)\}
\end{equation}
where $M$ stands for the number of unlabeled training samples, each associated with $N$ pseudo-labels and corresponding hidden states. In total, the augmented method outputs $M * N$ label-embedding pairs for further processing.

\subsection{Uncertainty Estimation}
Following \citet{li2005setred} and \citet{lang2022co}, we estimate the confidence of all pseudo-labels using the SETRED algorithm. The motivation of this algorithm is that training samples with similar embeddings are likely to have the same pseudo-labels. On the other hand, if a training sample is located near samples with different pseudo-labels in the embedding space, its own pseudo-label is likely to be uncertain. Using the output data-embedding-label set shown in Equation \ref{eq:c}, we can calculate the nearest neighbors of each training sample and estimate the labeling consistency.

To estimate the uncertainty of $y_u$, the pseudo-label of training sample $x_u$, we calculate the Euclidean distances between $x_u$ and all other $M*N -1$ samples using the calculated text embeddings. We construct a set of the top k nearest neighbors of $x_u$, namely $N(u)$. With the nearest neighbor set, an uncertain score of $(x_u, y_u)$ can be calculated as follows,
\begin{equation}
J_u = \sum_{v \in N(u)} \mathbb{I}(y_u \neq y_j) \;/\; (1 + \|e_u - e_v\|_2)
\end{equation}
where $\mathbb{I}$ is a binary indicator function, whose value is 1 when $y_u \neq y_v$ and 0 otherwise. $\|e_u - e_v\|_2$ stands for the Euclidean distance between the embeddings of $x_u$ and $x_v$. As a result, $J_u$ would have a higher value when more near neighbors are associated with different pseudo-labels.

To estimate the uncertainty of $y_u$, we compare $J_u$ with a null hypothesis where all pseudo-labels in $C$ except $y_u$ are randomly shuffled. After the shuffling, the entire data-label mapping set becomes uncertain. The expectation and variance of $J_u$ after shuffling is
\[
\mathbb{E}_v[J_u] = (1 - \hat{P}_{y_u}) \sum_{v \in N(u)}1 / (1 + \|e_u - e_v\|_2)
\]
\[
\sigma(J_u)^2 = \hat{P}_{y_u} \* (1 - \hat{P}_{y_u}) \* \sum_{v \in N(u)} 1 / (1 + \|e_u - e_v\|_2)^2
\]
The uncertainty can be estimated by verifying the significance of the difference between $J_u$ and the null hypothesis. An uncertainty score can be calculated as
\begin{equation}
s(u) = \frac{J_u - \mathbb{E}_v[J_u]}{\sigma(J_u)}
\end{equation}

With this method, we calculate uncertainty scores for all $M*N$ training samples in $C$ for further processing.

\subsection{Filtering and Relabeling}
After finishing estimating the uncertainty of each training sample, we sort all training samples in $C$ by their uncertainty scores and remove the 20\% most uncertain training samples. The remaining samples are used for relabeling based on majority voting. For example, a training sample $x_i$ has $N$ pseudo-labels $[y_0^i, y_1^i, \dots, y_{N-1}^i]$ after the augmented labeling step, and $n$ labels are removed based on the uncertainty scores.

The final pseudo-label of $x_i$ is decided by the voting result of the $N - n$ remaining labels. If all generated pseudo-labels of a training sample are removed or there is a tie in the voting, we re-run the labeling process without dropout to get the final pseudo-label. Following this approach, we keep all training samples and, meanwhile, obtain a more robust pseudo-label set.

\section{Experiments}
\label{sec:exp}
\noindent \textbf{Benchmarks.} We conduct experiments on popular natural language understanding tasks in the GLUE \cite{wang2018glue} benchmark, including RTE \cite{dagan2005pascal}, QNLI \cite{rajpurkar2016squad}, QQP, SST-2 \cite{socher2013recursive}, and CoLA \cite{warstadt2019neural}. We also assess the robustness of the proposed method against adversarial evaluation sets in the AdvGLUE corpus \cite{wang2021adversarial}, including Adv-QNLI, Adv-QQP, Adv-RTE, and Adv-SST2. The data in AdvGLUE is created by adding word-level and sentence-level perturbations to the GLUE data, as well as human-crafted examples. For Multi-Classification, we use Copa \cite{superglue} (which consists of questions paired with two answer choices), Emotion Classification \cite{CARER}, Amazon Review \cite{amazonreviews} and Ag-News \cite{agnews}. More details are shown in Appendix \ref{ap:data}.

\noindent \textbf{Hyper-parameters.} We train 350M RoBERTa \cite{devlin2018bert} and DeBERTa \cite{he2020deberta} models for the language understanding tasks, without using larger language models like GPT-3 \cite{brown2020language} or T0 \cite{sanh2021multitask} that are used for generating pseudo-labels in \cite{lang2022co}. We also use the same hyper-parameters across all tasks, attempting to avoid the problems mentioned in \citet{perez2021true}. In the entailment pretraining on the MNLI dataset \cite{N18-1101}, we optimize both RoBERTa and DeBERTa models with the AdamW optimizer \cite{loshchilov2018decoupled}. For all tasks and both models, we set $\varepsilon = 10^{-6}$. In the entailment pretraining, we set the weight decay weight to $10^{-5}$, and the learning rate for both models is 3e-6. During the self-training step, the learning rate of both models on all binary classification tasks is 4e-6 and is 1e-6 on multi-classification tasks, and the weight decay is constantly $10^{-2}$. We run the entailment pretraining for 2 epochs and the self-training for 6 epochs. In confidence-based labeling, we drop 1/8 data with the lowest confidence.

\noindent \textbf{Self-training details.} For each binary classification task, we randomly select $N = 2000$ unlabeled data examples. For each multi-classification task, we randomly select $N = 50$ unlabeled data examples. To estimate the uncertainty of the pseudo-labels in SETRED and SimPLE algorithms, we use the hidden states of the 4th layer from the top of both RoBERTa and DeBERTa language models as the supposition embeddings and measure the uncertainty with 9 neighbors. In SimPLE, we run 7 inferences for each training sample with different dropouts. We train and evaluate the models for each task with 10 independent runs on 2 V100 32G GPUs. Each experiment takes less than an hour. 

\noindent \textbf{Assessment.} We evaluate the performance of our algorithm by comparing the average classification accuracy against baseline methods and the robustness. We describe the term \emph{Robustness} as follows: in multiple independent experiments, a robust method should achieve high maximum, minimum, and average accuracy against with different backbone model and training data, on different natural language understanding tasks. 

\subsection{GLUE and AdvGLUE Tasks}

\begin{table*}[t]
\vspace{-4mm}
\centering
\small
\begin{tabular}{@{}llllllllllll@{}}
\toprule
\multirow{2}{*}{\textbf{Method}} & \multicolumn{5}{c}{\textbf{GLUE}}                          & \multirow{2}{*}{\textbf{Method}} & \multicolumn{5}{c}{\textbf{AdvGLUE}}                       \\ \cmidrule(lr){2-6} \cmidrule(l){8-12} 
      & \multicolumn{1}{c}{\textbf{QNLI}} & \multicolumn{1}{c}{\textbf{QQP}} & \multicolumn{1}{c}{\textbf{RTE}} & \multicolumn{1}{c}{\textbf{SST2}} & \multicolumn{1}{c}{\textbf{Avg.}} &       & \multicolumn{1}{c}{\textbf{QNLI}} & \multicolumn{1}{c}{\textbf{QQP}} & \multicolumn{1}{c}{\textbf{RTE}} & \multicolumn{1}{c}{\textbf{SST2}} & \multicolumn{1}{c}{\textbf{Avg.}} \\ \midrule
\multicolumn{12}{c}{Few-shot (left) and fully-supervised (right) medium LMs (350M) with human-generated labels}                               \\ \hdashline[1.5pt/2pt]
PET                              & 61.3   & 67.6  & 65.7  & 91.8 & 71.6  & R3F   & 47.5   & 40.6  & 50.1  & 38.5 & 44.2  \\
LM-BFF                           & 69.2   & 69.8  & 83.9  & 90.3 & 78.3  & CT$_T$                            & 49.6   & 40.7  & 46.2  & 39.2 & 43.9  \\
P-tuning                         & 58.8   & 67.6  & 70.8  & 92.6 & 72.5 & MT    & 47.5   & 41.5  & 52.5  & 51.3 & 48.2  \\
PPT & 68.8 & 67.2 & 67.9 & 92.3 & 74.1 & BERT & 39.8 & 37.9 & 40.5 & 33.0 & 37.8 \\
UPT & 70.1 & 72.1 & 68.9 & 92.9 & 76.0 & RoBERTa & 52.5 & 45.4 & 62.8 & 58.5 & 54.8 \\
EFL & 68.0 & 67.3 & \textbf{85.8} & 90.8 & 78.0 & DeBERTa & 57.9 & 60.4 & \textbf{79.0} & 57.8 & 63.8 \\
\midrule
\multicolumn{12}{c}{Few-shot large LMs (137B) with human-generated labels}                                   \\ \hdashline[1.5pt/2pt]
LaMDA                            & 55.7   & 58.9  & 70.8  & 92.3 & 69.4 & \textbackslash{}                 & -      & -     & -     & - & -     \\
FLAN                             & 63.3   & 75.9  & 84.5  & \textbf{94.6} & 79.6 & \textbackslash{}                 & -      & -     & -     & - & -     \\ \midrule
\multicolumn{12}{c}{Zero-shot adaptation of entailment classifiers based on medium LMs (350M)}                                   \\ \hdashline[1.5pt/2pt]
DeBERTa-Cat                       & 71.6   & 70.5  & 74.0  & 84.6 & 75.2  & \textbackslash{}                 & 60.8      & 47.4     & 50.6     & 56.1 & 53.7    \\
RoBERTa-Sup                       & 71.5   & 78.6  & 81.2  & 87.7 & 79.8  & \textbackslash{}                 & 62.1      & 52.6     & 61.7     & \textbf{59.9} & 59.1    \\
DeBERTa-Sup  & 77.3               & \textbf79.9  & 84.5  & 90.1 & 82.9 & \textbackslash{}                 & 61.5      & 64.1     & 66.7     & 42.6 & 58.7    \\ \midrule
\multicolumn{12}{c}{Self-trained \texttt{RoBERTa-large} (350M) without human-generated labels}                  \\ \hdashline[1.5pt/2pt]
Baseline-ST                       & 74.1   & 80.1  & 81.5  & 88.3 & 81.0 & Baseline-ST                 & 64.9   & 60.6  & 60.9  & 56.6 & 60.8 \\
Dropout                      & 78.5   & 80.5  & 80.9  & 88.8 & 82.2 & Dropout                 & 69.2   & 57.8  & 61.9  & 57.3 & 61.6 \\
SETRED                       & 80.5   & 80.5  & 80.8  & 88.3 & 82.5 & SETRED                 & 68.0   & 56.5  & 62.6  & 58.9 & 61.5 \\
SimPLE (ours)                      & 83.1   & 80.7  & 83.1  & 91.6 & 84.6 & SimPLE (ours)                 & 69.6   & 54.4  & 62.3  & 58.8 & 61.3 \\ \midrule
\multicolumn{12}{c}{Self-trained \texttt{DEBERTa-large} (350M) without human-generated labels}                                           \\ \hdashline[1.5pt/2pt]
Baseline-ST                       & 79.0   & 80.2  & 83.4  & 92.1 & 83.7 & Baseline-ST                 & 65.8   & \textbf{70.4}  & 68.4  & 50.9 & 63.9 \\
Dropout                      & 81.1   & 80.5  & 84.1  & 91.8 & 84.4 & Dropout                 & 70.1   & 63.3  & 70.9  & 49.9 & 63.6 \\
SETRED                       & 83.4   & 80.5  & 83.9  & 92.0 & 84.9 & SETRED                 & 69.8   & 69.5  &69.9  & 50.9 & 65.0 \\
SimPLE (ours)                      & \textbf{85.2}  & \textbf{81.0}  & 85.5 & 92.8 & \textbf{86.1}  & SimPLE (ours) & \textbf{70.1} & 68.1  & 73.8 & 51.6 & \textbf{65.9}  \\ \bottomrule
\end{tabular}
\caption{Experimental results on binary classification tasks with 10 independent experiments. Cat stands for concatenation-based pretraining and Sup stands for supposition classification.}
\label{tab:mean-results}
\end{table*}

\begin{table}[h]
\centering
\small
\begin{tabular}{@{}llllll@{}}
\toprule
\multirow{2}{*}{\textbf{Method}} & \multicolumn{5}{c}{\textbf{Multi-Classification}}               \\ \cmidrule(lr){2-6} 
      & \textbf{Copa} & \textbf{EM} & \textbf{AR} & \textbf{News} & \textbf{Avg} \\ \midrule
\multicolumn{6}{c}{ \texttt{DEBERTa-large} (350M)}                  \\ \hdashline[1.5pt/2pt]
Pretrain      & 77.0 & 51.93 & 37.01 & 73.40 & 59.84\\
BaseST                       & 78.75   & 51.24  & 38.80  & 73.10 &60.47 \\
Dropout                      & 78.25   & 53.69  & 38.19  & 73.16 &60.82 \\
SETRED                       & 78.0   & 52.42  & 37.61  &  73.33 &60.34\\
SimPLE               & \textbf{79.75}   & \textbf{54.58}  & \textbf{39.05}  & \textbf{73.57} &\textbf{61.74} \\ \midrule
\multicolumn{6}{c}{ \texttt{RoBERTa-large} (350M)}                                           \\ \hdashline[1.5pt/2pt]
Pretrain      & 76.0 & 49.21 & 33.31 & 63.18 &55.43\\
BaseST                       & 76.67   & 50.94  & 37.38  & 64.64 & 57.41 \\
Dropout                      & 78.67   & 50.99  & 42.87  &  61.05 & 58.40 \\
SETRED                       & 78.0   & 50.53  & 27.16  &  63.24 & 54.73\\
SimPLE                      & \textbf{79.0}  & \textbf{51.79}  & \textbf{44.06} & \textbf{65.60} & \textbf{60.11} \\ \midrule
\multicolumn{6}{c}{ \textit{Large Language Models}} \\ \hdashline[1.5pt/2pt]
Zero-shot & 70.0$^{\diamondsuit}$ & 42.7$^{\ddag}$ & - & 43.9$^{\ddag}$ & - \\
Few-shot  & 77.0$^{\dag}$ & - & - & 61.0$^{\ddag}$ & -  \\
\midrule
Class Num & 2 & 6 & 5 & 4 &- \\
\bottomrule
\end{tabular}
\caption{Multi-class NLU results with 3 independent runs. The Copa model selects from two candidate sentences, which is different from the previous binary NLU tasks. $\diamondsuit$: \texttt{T5-11b}, $\dag$: \texttt{GPT-Neo-6b}, $\ddag$: \texttt{GPT-3-175b}. The performance of large language models are cited from \citet{zhao2021calibrate} and \citet{wang2023large}.}
\vspace{-4mm}
\label{tab:multi-class}
\end{table}

The experiment results are shown in Table \ref{tab:mean-results}. We compare the adaptation performance of entailment-based language models and the improvement of different self-training approaches.

\noindent \textbf{Compare with supervised baselines.} We compare our entailment self-training methods with few-shot fine-tuning baselines. The few-shot baselines, including PET \cite{schick2021exploiting}, LM-BFF \cite{gao2020making}, P-tuning \cite{liu2021gpt}, PPT \cite{gu2021ppt}, and UPT \cite{wang2022towards}, are based on 350M BERT or RoBERTa backbones. 
Our pretrained DeBERTa entailment model outperforms the best few-shot baseline (LM-BFF) by 4.5\%, and the RoBERTa entailment model outperforms LM-BFF by 1.5\%. With self-training, our SimPLE method further improves the model's performance by a large margin. The RoBERTa performance is boosted by nearly 5\% and the average performance of DeBERTa is over 86\%, outperforming the best few-shot supervised baselines by 6.9\%.

On the other hand, we compare our model with fully supervised RoBERTa/DeBERTa models and robust training methods, including R3F \cite{aghajanyan2020better}, child tuning (CT) \cite{xu2021raise}, and match tuning (MT) \cite{tong2022robust} models, on the AdvGLUE benchmark. We found that the fully-supervised DeBERTa model is the best baseline on the AdvGLUE benchmark. However, our RoBERTa entailment model outperforms all robust training baselines with the same pretrained backbone by over 10\%. With SimPLE self-training, the DeBERTa entailment model achieves the best performance on AdvGLUE, outperforming the fully-supervised DeBERTa model by 2.1\% as well as all other baselines.

We found that our pretrained entailment models outperform EFL, the few-shot fine-tuned entailment model based on \texttt{RoBERTa-large} proposed by \citet{wang2021entailment}. The self-trained models further outperform EFL with larger margins. This indicates the strong adaptation ability introduced by the supposition-based NLU strategy.

\noindent \textbf{Compare with large language models.} We found that both zero-shot pretrained and semi-supervised self-trained entailment models outperform the few-shot large language models on QNLI, QQP, and RTE tasks, and achieve significantly higher average accuracy on GLUE. This suggests that our method is computation-efficient - the models use 1/400 parameters, without human-generated task-specific labels, but achieve better performance than expensive large-scale language models on NLU tasks.

\begin{figure*}[t]
\centering
\includegraphics[width=\textwidth]{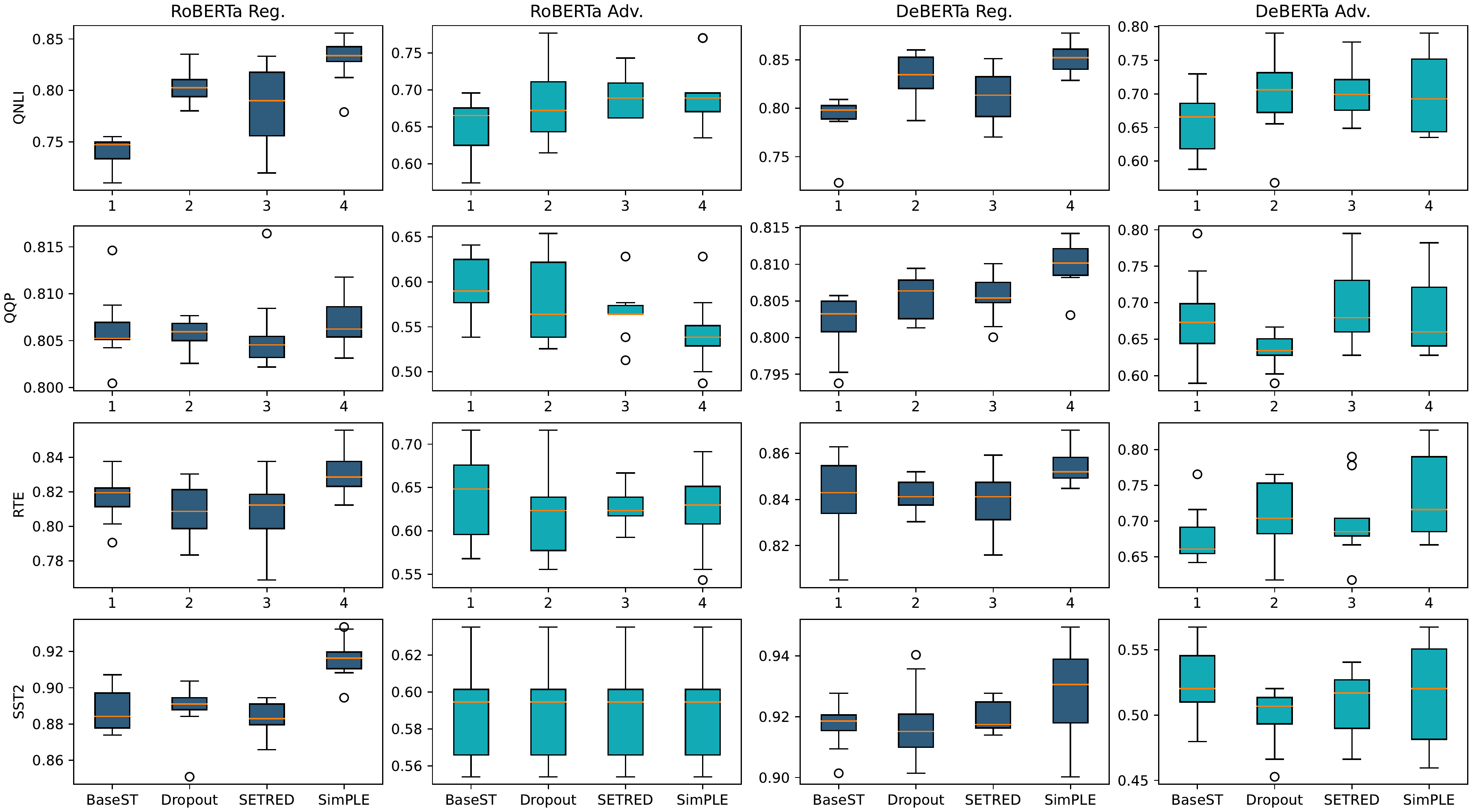}
\caption{The results of 10 independent experiments with self-trained RoBERTa and DeBERTa models on GLUE (*BERTa Reg.) and AdvGLUE (*BERTa Adv.) benchmarks.}
\label{fig:robustness}
\end{figure*}

\noindent \textbf{Compare with self-training baselines.} By averaging 10 independent evaluations across GLUE and AdvGLUE benchmarks and backbone models, we found that Dropout and SETRED improve baseline self-training performance on a similar level. On average, SETRED outperforms Dropout by 0.5\% on 4 experiment settings. On the GLUE benchmark, the SimPLE method improves the model's performance by 1.5 to 2\% on average. The highest improvement boost is on the QNLI tasks, where the SimPLE self-training method outperforms the baseline self-training by 9\% and 6\% on RoBERTa and DeBERTa respectively. Although the average improvement is not very high, we will show that SimPLE is significantly more robust. The results show that augmenting the pseudo-labels without removing uncertain training samples benefits self-training, which aligns with our hypothesis.

In general, the experiments on binary classification NLU tasks proved the data and computation efficiency of entailment self-training over different strong baseline models. Furthermore, the SimPLE algorithm we propose in this work achieves the best average performance, significantly outperforms all baselines on some of the tasks, and meanwhile preserves the robustness of entailment models against adversarial benchmarks.

\subsection{Multi-class NLU Tasks}
The experiment results on Copa, Emotion, Amazon Review, and Ag News are shown in Table 3. In multi-classification tasks, we present the comparative results of the pretrained entailment-based language models and the 4 self-training approaches compared in the previous section with binary NLU tasks, including standard self-training, dropout-based re-labeling, SETRED, and SimPLE.

\noindent \textbf{The effect of dropout-based augmentation.} 
By merely using dropout, the augmented self-training outperforms the standard normal self-training baseline which keeps all the pseudo-labels in general. This further validates the previous finding that by adding dropout, the models adopt noises that benefit the inference, generate augmented pseudo-labels and mitigate the overfitting problem.

\noindent \textbf{The effect of SETRED.}
By merely using SETRED, the self-training does not see a consistent improvement in performance and even falls behind the pretrained and standard self-trained models that preserve all pseudo labels in some tasks (like Amazon-Review). This fact suggests that removing uncertain pseudo-labels can lead the model to overfit confident training samples, thus hurting the self-fine-tuning performance.

\noindent \textbf{The effect of SimPLE.}
Table \ref{tab:multi-class} shows that the SimpLE algorithm constantly outperforms all pretrained and self-trained baselines on both backbone models across all multi-class benchmarks, which aligns with the result on the binary NLU tasks. This fact further validates our hypothesis that augmenting the pseudo-labels of uncertain training samples can improve the performance of self-training.

\noindent \textbf{Compare with Large Language Models.} We notice that our self-trained methods can outperform several large language models. On Emotion and AG News tasks, the pretrained entailment model without self-training can achieve a significant improvement over the \texttt{GPT-3-175b} model, which is 500 times large than the entailment model. This indicates that the entailment-based model is a more efficient and trustworthy option for many natural language understanding tasks.

\section{Analysis}
\label{sec:analysis}
\noindent \textbf{Robustness.} Besides the mean accuracy of all experiments, we also visualize the results of all independent evaluations of different self-training strategies in Figure \ref{fig:robustness}. We found that SimPLE constantly outperforms other self-training baselines on the regular GLUE benchmark by comparing mean, maximum, and minimum accuracy. Although DeBERTa performs similarly under different self-training strategies on QQP in terms of average accuracy, there exists a significant gap between the minimal performance of baseline and SimPLE. This indicates that SimPLE is more robust and safer compared with the regular self-training algorithm. The only exception is the DeBERTa model on SST2 - the mean performance of SimPLE is better, but it has a lower minimal performance than the baseline self-training method.

Most models overfit to the training corpora and achieve high accuracy on regular evaluation sets, but perform poorly on adversarial benchmarks \cite{wang2021adversarial}. As a result, fully supervised models achieve less than 60\% accuracy on AdvGLUE. We also investigate if SimPLE hurts the model's robustness against adversarial evaluation data. We found that, except RoBERTa on AdvQQP, other settings show that the entailment-based models are still robust after SimPLE self-training. As we compared in Table \ref{tab:mean-results}, all these results significantly outperform fully-supervised baselines.

\begin{figure}[t]
\centering
\includegraphics[width=.8\columnwidth]{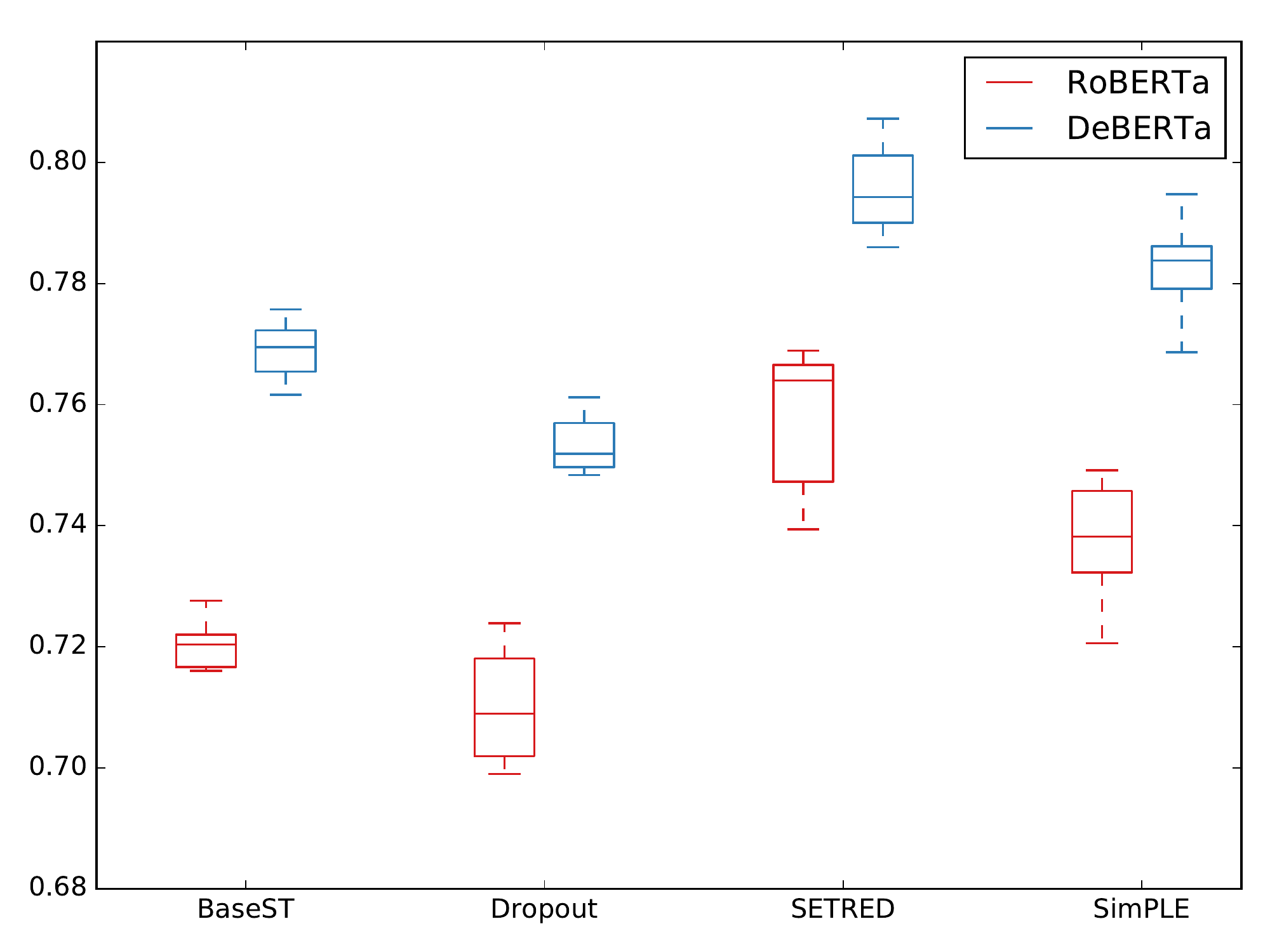}
\caption{Pseudo-labeling accuracy of entailment models with standard (ST), dropout, SETRED, and SimPLE strategies. SETRED achieves higher accuracy because uncertain data samples are dropped.}
\label{fig:placc}
\end{figure}

\noindent \textbf{Pseudo-labeling Accuracy.} We show the pseudo-labeling accuracy of RoBERTa and DeBERTa-based entailment models with different strategies in Figure \ref{fig:placc} with 10 independent experiments. The results indicate that the DeBERTa models predict more accurate pseudo-labels in general. On the other hand, the pseudo-label sets produced by SimPLE with both models are significantly less noisy than the standard and dropout-based labeling methods without removing any uncertain data samples. SETRED achieves the highest labeling accuracy because it drops uncertain samples. The comparison suggests that SimPLE achieves the highest performance because it achieves high pseudo-labeling accuracy on uncertain training samples.

\begin{figure}[t]
\centering
\includegraphics[width=.9\columnwidth]{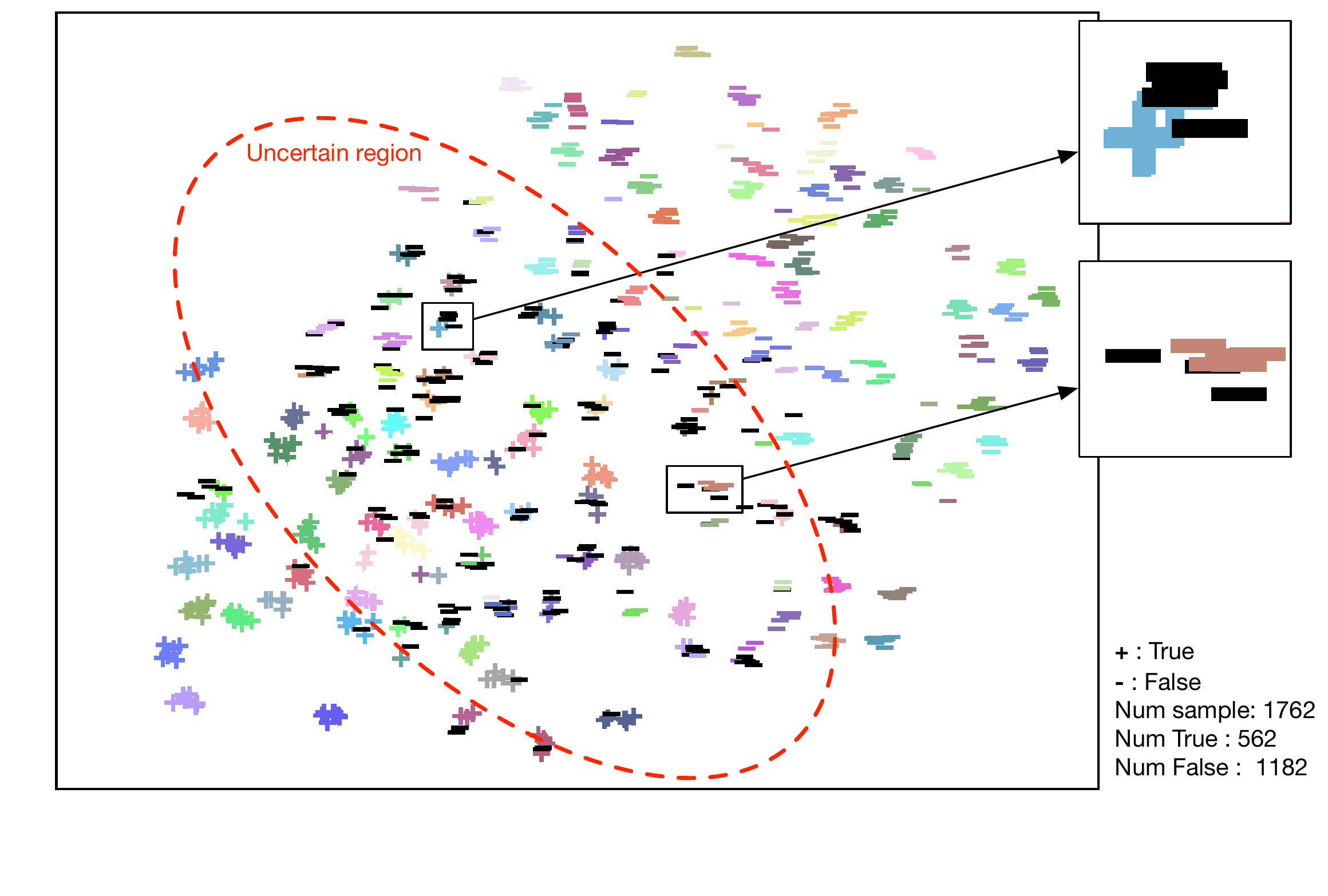}
\caption{Visualization of the embeddings, pseudo-labels, and uncertainty of QNLI suppositions calculated by the pretrained DeBERTa entailment model. Each data sample has 7 embeddings calculated with different dropouts. Black stands for uncertain points and other colors stand for different training examples.}
\label{fig:emb}
\end{figure}

\noindent \textbf{Case study.} We visualize the hidden states, pseudo-labels, and confidence of the training samples in the QNLI tasks calculated by the pretrained DeBERTa entailment model with the SimPLE algorithm in Figure \ref{fig:emb}. The embedding space is calculated with t-SNE \cite{van2008visualizing} using 252 training samples with 252*7=1764 embeddings. Half of them are plotted in the figure. Each training sample is evaluated with 7 different dropouts, and the uncertainty is estimated with 9 neighbors. In Figure \ref{fig:emb}, different embeddings of the same training sample are labeled with the same color, while the uncertain cases are marked in black. \textbf{+} and \textbf{-} stand for the truth value of the suppositions. As shown in the figure, most uncertain cases appear around the uncertain circle. We also highlight two training samples with uncertain representations. This phenomenon indicates that the SimPLE algorithm can drop most embeddings of a data sample and edit the voting results of the dropout-based pseudo-labeling method, improving the pseudo-labeling accuracy from 76.5\% to 79.2\% in this experiment.

We also show that the original pseudo-label set is unbalanced, with 67.1\% of all predicted labels being ``False''. Although we do not provide any prior knowledge about the label distribution of the task (unknown without human annotation), the SimPLE method mitigates the bias through the uncertain candidate removal process. Figure \ref{fig:emb} shows that most uncertain pseudo-labels estimated by SimPLE are ``False'', thus the remaining pseudo-labels are more balanced.

\section{Conclusion}
We show that entailment-based language models can be adapted to different NLU tasks without supervision and achieve robust performance against noisy pseudo-labels and adversarial texts. We design a supposition-based prompting strategy to improve the zero-shot adaptation performance of entailment-based models. To improve the stability of self-training, we propose the SimPLE algorithm for augmented pseudo-labeling. Experiments on binary, multi-class, regular, and adversarial NLU tasks show that the SimPLE self-training strategy significantly outperforms a number of strong baselines, including 400 and 500 times larger language models on both zero-shot and weakly supervised settings, proving the effectivenss of entailment self-training for efficient and trustworthy natural language understanding systems.

\section*{Limitations}
Our method utilized pretrained entailed models and adapted them to other domains under zero-shot and self-training settings. There are two limitations that we would like to improve in future work. Firstly, we use human-designed suppositions for each task, which is less automatic than a direct, zero-shot adaptation of the models. Secondly, the self-training on some multi-class classification tasks is not as high as on binary NLU tasks, indicating the challenge of applying entailment models to multi-choice tasks. We would like to overcome this in the next step.

\section*{Ethics Statement}
We propose a method that can significantly reduce the financial and environmental cost of language model learning. By reducing the need for data collection and human labeling, our method can effectively protect user and data privacy by avoiding leaking any information while building the training corpora. We found that a medium-sized language model can achieve similar performance as the state-of-the-art large-scale language models, suggesting that we can cost less financially and environmentally during model training and evaluation for comparable performance. However, since we reduced the need for human-labeling efforts, the deployment of the system might decrease the number of data annotation jobs.

\bibliography{anthology,custom}
\bibliographystyle{acl_natbib}

\appendix
\section{Data Details}
\label{ap:data}
\noindent \textbf{GLUE/AdvGLUE} In this work, we evaluate our method with the GLUE\footnote{\url{https://gluebenchmark.com/}} and AdvGLUE\footnote{\url{https://adversarialglue.github.io/}} benchmarks. We pretrain our models on MNLI, and evaluate on all other AdvGLUE tasks, AdvQNLI, AdvQQP, AdvRTE, and AdvSST2. we also evaluate the models on the regular versions of these tasks in GLUE. The statistics of the GLUE and AdvGLUE benchmarks are shown in Table \ref{tab:data_details}.

\begin{table}[h]
\centering
\begin{tabular}{@{}lccc@{}}
\toprule
\textbf{Corpus} & \multicolumn{1}{c}{\textbf{|Train|}} & \multicolumn{1}{c}{\textbf{|Test|}} & \multicolumn{1}{c}{\textbf{|Adv-Test|}} \\ \midrule
MNLI            & 393k                                 & 20k                                 & 1.8k                                    \\
QNLI            & 105k                                 & 5.4k                                & 0.9k                                    \\
QQP             & 364k                                 & 391k                                & 0.4k                                    \\
RTE             & 2.5k                                 & 3k                                  & 0.3k                                    \\
SST2            & 67k                                  & 1.8k                                & 1.4k                                    \\
\bottomrule
\end{tabular}
\caption{Statistics of the corpora used in this work}
\label{tab:data_details}
\end{table}

\noindent \textbf{Multi-Classification} In multi-class classification tasks, we evaluate our method with SuperGlue Copa\cite{superglue}, Emotion Classification\cite{CARER}, and Amazon Review\cite{amazonreviews}. The statistics of these corpora are shown in Table \ref{tab:multi_data_details}.  
\begin{table}[h]
\centering
\begin{tabular}{@{}lccc@{}}
\toprule
\textbf{Corpus} & \multicolumn{1}{c}{\textbf{|Train|}} & \multicolumn{1}{c}{\textbf{|Test|}} & \multicolumn{1}{c}{\textbf{|Class Num|}} \\ \midrule
Copa           & 400                                 & 100                                 & 2                                    \\
Emotion            & 16k                                 & 2k                                & 6                                    \\
AR             & 200k                                 & 5k                                & 5                                   \\
News  & 120k & 7.6k & 4 \\
\bottomrule
\end{tabular}
\caption{Statistics of the corpora used in multi-class classification}
\label{tab:multi_data_details}
\end{table}

\section*{Reproducibility}
\noindent \textbf{Data.} We introduce the tasks and corpora we used for training and evaluation in Section \ref{sec:exp} and Appendix \ref{ap:data}.

\noindent \textbf{Method.} We introduce the difference between our method and previous work in Section \ref{sec:bg}, the details of our method in Section \ref{sec:entst} and \ref{sec:simple}.

\noindent \textbf{Hyper-parameter.} We describe the key hyper-parameters of self-training in Section \ref{sec:exp}.

\noindent \textbf{Experiments.} We describe the experiment results, and number of independent runs in Section \ref{sec:exp}, and Section \ref{sec:analysis} to prove the statistical significance.


\end{document}